\documentclass[10pt,twocolumn,letterpaper]{article}

\usepackage[pagenumbers]{wacv} 

\usepackage{graphicx}
\usepackage{amsmath}
\usepackage{amssymb}
\usepackage{booktabs}

\usepackage{multirow}
\usepackage{subcaption}
\usepackage{array}
\usepackage{caption}
\usepackage{arydshln}
\usepackage{float}
\usepackage{paralist}
\usepackage{xcolor}

\usepackage[pagebackref,breaklinks,colorlinks]{hyperref}

\usepackage[capitalize]{cleveref}
\crefname{section}{Sec.}{Secs.}
\Crefname{section}{Section}{Sections}
\Crefname{table}{Table}{Tables}
\crefname{table}{Tab.}{Tabs.}

\def\wacvPaperID{17} 
\def\confName{WACV}
\def\confYear{2024}

\begin{document}

\title{HashReID: Dynamic Network with Binary Codes for Efficient Person Re-identification}

\author{Kshitij Nikhal$^{1,*}$
\and
Yujunrong Ma$^2$
\and
Shuvra S. Bhattacharyya$^2$
\and
Benjamin S. Riggan$^{1,*}$
\vspace{0.1mm}
\and
$^1$University of Nebraska-Lincoln, 1400 R St, Lincoln, NE 68588
\and
$^2$University of Maryland, College Park, 8223 Paint Branch Dr, College Park, MD 20742
}

\maketitle

\begin{abstract}
   Biometric applications, such as person re-identification (ReID), are often deployed on energy constrained devices. While recent ReID methods prioritize high retrieval performance, they often come with large computational costs and high search time, rendering them less practical in real-world settings. In this work, we propose an input-adaptive network with multiple exit blocks, that can terminate computation early if the retrieval is straightforward or noisy, saving a lot of computation. To assess the complexity of the input, we introduce a temporal-based classifier driven by a new training strategy. 
   Furthermore, we adopt a binary hash code generation approach instead of relying on continuous-valued features, which significantly improves the search process by a factor of 20. To ensure similarity preservation, we utilize a new ranking regularizer that bridges the gap between continuous and binary features. Extensive analysis of our proposed method is conducted on three datasets: Market1501, MSMT17 (Multi-Scene Multi-Time), and the BGC1 (BRIAR Government Collection). Using our approach, more than 70\% of the samples with compact hash codes exit early on the Market1501 dataset, saving 80\% of the networks computational cost and improving over other hash-based methods by 60\%. These results demonstrate a significant improvement over dynamic networks and showcase comparable accuracy performance to conventional ReID methods. Code will be made available. 
\end{abstract}
\vspace{-0.8cm}
\section{Introduction}
\label{sec:intro}

\begin{figure}[htb]
    \centering
    \includegraphics[width=\columnwidth]{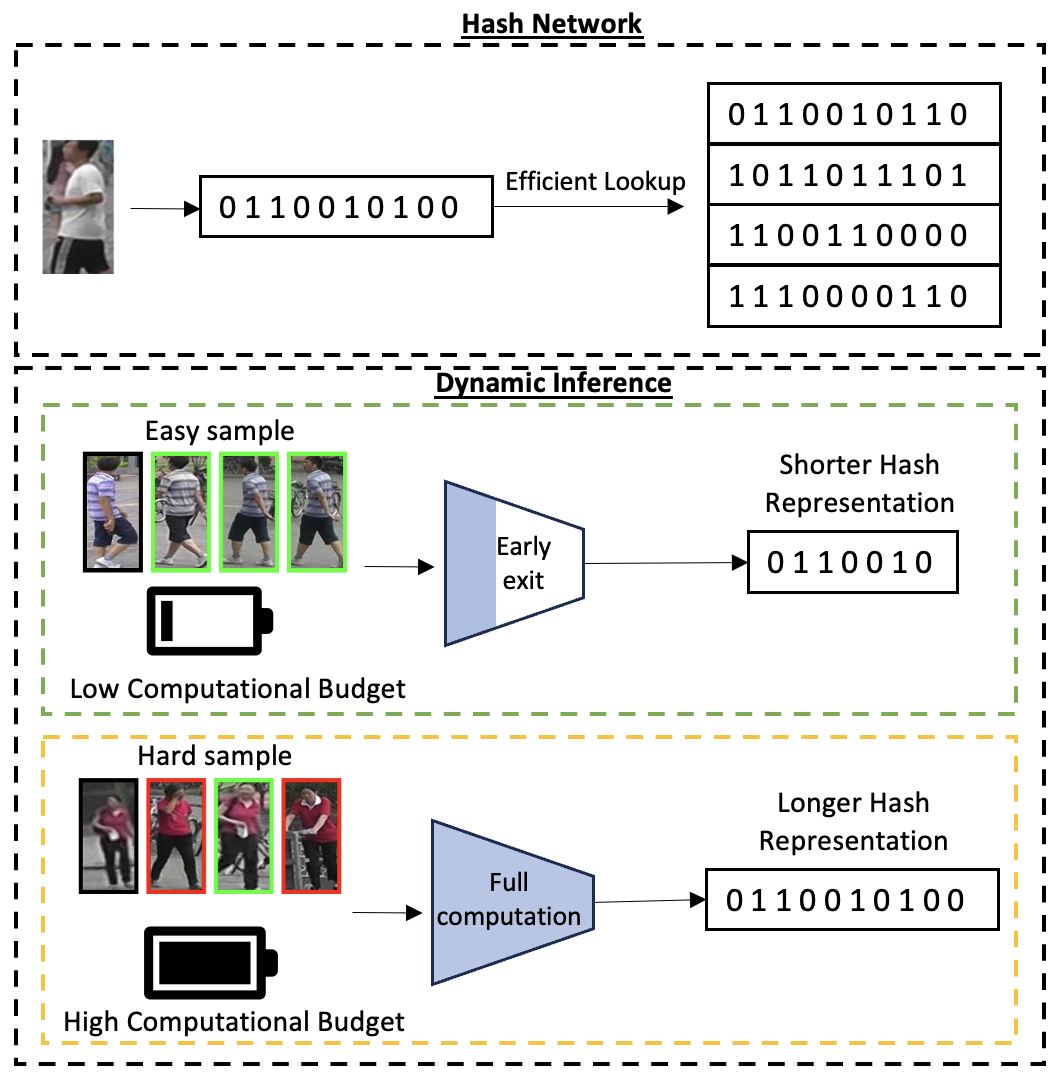}
    \caption{Overview of our contributions. Our method generates hash/binary representation of the features, to enable fast lookup. Moreover, the network adjusts its inference process according to the input saving computational costs. }
    \label{fig:introduction}
\end{figure}

Person re-identification (ReID), where probe (query) images are matched against gallery images is an important application in real-world scenarios. For example, unmanned aerial vehicles (UAVs) such as drones are often equipped with identification capabilities for applications such as border control, intelligence, and security. This task poses a significant challenge due to the considerable variations in factors such as pose, clothing, resolutions, occlusions, camera-viewpoints and more. Additionally, the deployment of ReID models in realistic scenarios, such as real-time ReID, is hindered by significant latencies induced by model complexity and searching (sorting).

To address this, researchers have designed efficient model architectures, such as incorporating depth-wise convolutions~\cite{howard2017mobilenets}, compound scaling for balancing width and depth~\cite{tan2019efficientnet}, point-wise group convolutions~\cite{zhang2018shufflenet}, and squeeze and expand layers~\cite{iandola2016squeezenet}. However, the high-dimensional representations still render the matching process inefficient.

Binary (hash) representations---the transformation of high-dimensional continuous-valued representations to discrete binary codes---have been recently used to accelerate the matching process~\cite{zhao2018deepssh, wang2020faster, Cao_2017_ICCV, liu2019adversarial}. For instance, comparing two 2048-dimensional representations with Hamming distance metric is 229x faster compared to continuous valued features~\cite{wang2020faster}. However, the computational time and energy for network inference still remains a bottleneck. 

Another recent approach focuses on adaptive inference, where the network dynamically adjust its architecture based on the complexity of the input~\cite{teerapittayanon2016branchynet, wang2018skipnet}. 
For example, ElasticNet~\cite{zhou2019elastic, hector2021scalable} generates intermediate outputs and jointly optimizes the loss of all layers, allowing termination of computation to meet changing computational demands.
However, spatial information at the earlier layer is often not correctly utilized, leading to subpar initial performance.

In this work, we explore the combination of both hash representations and dynamic inference (see Figure~\ref{fig:introduction}), achieving performance competitive with traditional neural networks. We argue that most of the discriminability is lost in the earlier layers due to the global pooling on the large spatial dimensions, and instead utilize part-based local pooling to boost performance. Due to the need for extensive fine-tuning of threshold-only methods used to determine when to stop computation, we opt for the utilization of a learnable exit policy to make predictions. Finally, we make the hash learning tractable and discriminative by employing a soft-sign operation driven by a ranking regularizer to preserve the similarity between the continuous-valued and binary discrete-valued features. 

Our work does not rely on the underlying network architecture and can be adapted to various encoders~\cite{howard2017mobilenets, iandola2016squeezenet, tan2019efficientnet}, providing an efficient way to dynamically terminate computation during inference. 
Our contributions can be summarized as follows:\\
    1. We propose a novel hash-based network called HashReID, which leverages spatial information in earlier layers to generate robust representations at early exit points, while also generating a compact hash representation for efficient inference and lookup. \\
    2. We introduce a new ranking regularizer that maintains the similarity between continuous and binary features. \\
    3. We present a novel policy called Exit using Training Statistics (ETS) that uses a gated recurrent unit (GRU) to train and predict the difficulty of samples as easy, hard or impossible to recognize.  

We conduct extensive analysis on the Market1501~\cite{zheng2015scalable}, MSMT17~\cite{wei2018person}, and the BGC1~\cite{cornett2023expanding} datasets, and demonstrate competitive performance in  realistic situations, such as budgeted performance metrics.

\begin{figure*}[htb]
     \centering
     \includegraphics[width=2\columnwidth]{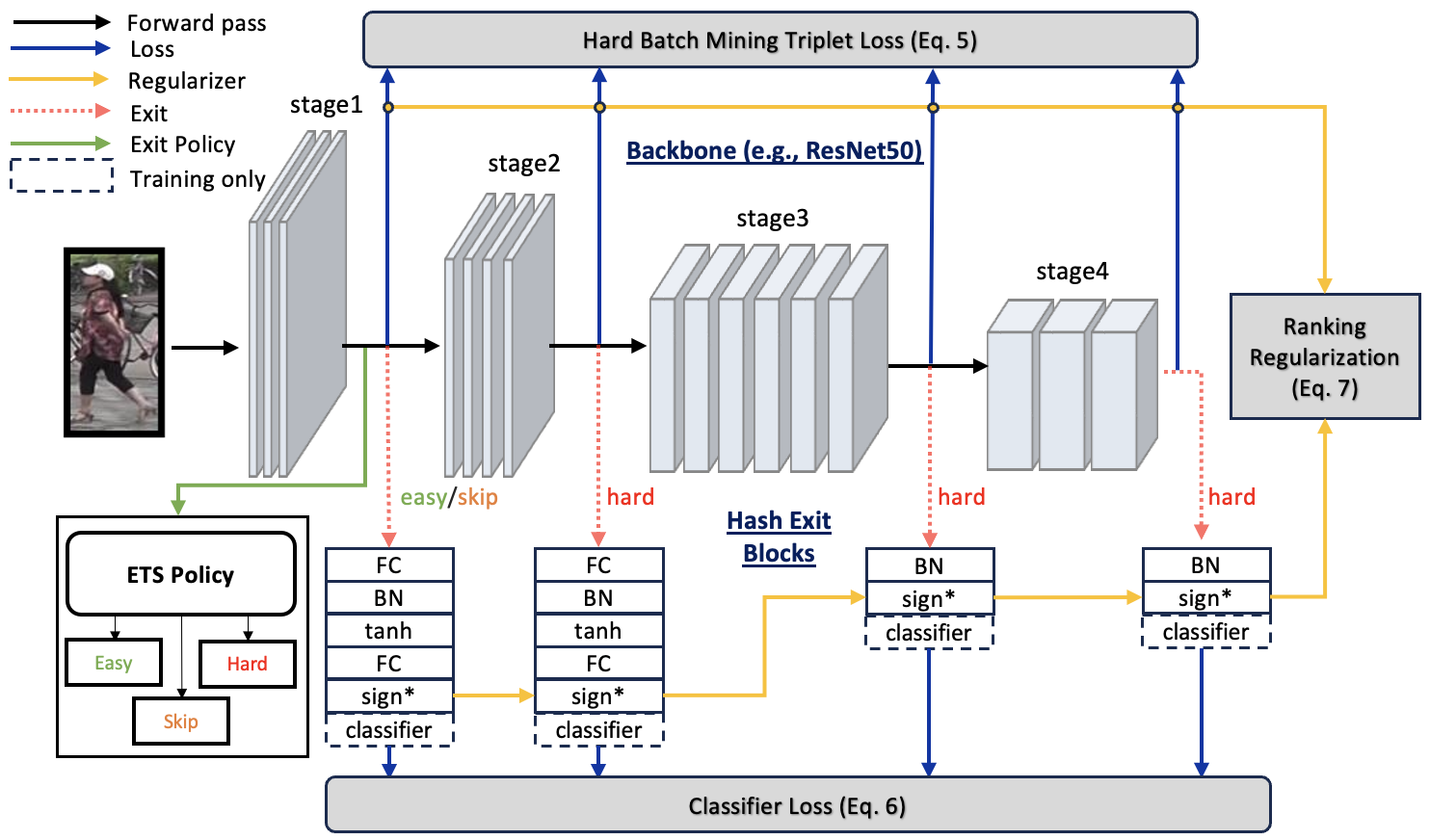}
     \caption{Our proposed method uses intermediate feature representations, driven by the triplet loss as specified in Eq~\ref{eq:triplet}, as input to the hash exit blocks that is optimized using the classifier loss. The stage-1 part-based pooled features ($\hat{s_1}$) is used by the ETS policy to determine the layer at which to exit. The easy/skip predictions are exited at the initial layer, while the hard prediction samples are exited using the query-gallery margin heuristic. Sign* denotes soft-sign activation used during training, and sign activation during the inference phase. }
     \label{fig:methodology}
 \end{figure*}

\section{Related Work}
\label{sec:related}

\textbf{Supervised ReID:}
Supervised ReID approaches have seen tremendous progress, especially with the exploration of the triplet loss for ReID~\cite{hermans2017defense}. To tackle misaligned person crops, HaCNN~\cite{li2018harmonious} proposes a local and global branches that jointly learns soft and hard attention to focus on discriminative regions. OSNet~\cite{zhou2019omni} fuses feature representations from multiple feature scales within and across channels, thereby generating a `omni-scale' feature representation. BOT~\cite{luo2019bag} summarizes training strategies most effective for person ReID. The challenge of deploying these methods lies in the requirement for efficient computation and inference, particularly in real-time applications. On the other hand, our approach enables early computation termination while simultaneously generating hash representations, thereby accelerating the matching process.

\textbf{Adaptive Networks:}
As input samples vary in difficulty for recognition 
 or classification tasks, a recent paradigm has emerged to skip layers or intermediately exit the network prediction to save computation and energy costs. BranchyNet~\cite{teerapittayanon2016branchynet} adds two early-exit branches having a combination of $3 \times 3$ convolutional layers and fully connected layers at equidistant locations. 
 ElasticNet~\cite{zhou2019elastic, hector2021scalable} inserts exit pathways following each
residual block, resulting in a total of 17 exits for the ResNet50~\cite{he2016deep} model.
 MSDNet~\cite{huang2017multi} uses dense multi-scale features to learn intermediate classifiers and inputs are exited once a confidence threshold is reached. RANet~\cite{yang2020resolution}  improves upon MSDNet by conditioning on the resolution of the input sample by utilizing sub-networks with different input resolutions. In DareNet~\cite{wang2018resource}, a multi-resolution approach is applied on the ReID task by inserting early exit blocks after every stage of the ResNet50 network. In contrast to existing methods, our approach takes advantage of spatial information in the early layers of the network to achieve competitive performance at an early stage. Additionally, we introduce a novel exit policy and generate a hash representation of the feature vector, resulting in a significant acceleration of query lookup.

\textbf{Hashing Networks:}
For retrieval problems, hash representations have been explored because of their capability of efficient lookup and storage. Deep learning methods suffer from the ill-posed gradient problem when learning a binary code due to the discontinuity of the signum (sign) function at zero. To address this, HashNet~\cite{Cao_2017_ICCV} begins learning using a hyperbolic tangent (tanh) activation, and gradually modifies to approximate the sign activation.  
 Kernel-Based Supervised Hashing (KSH)~\cite{6247912} learns a kernel to map data to binary codes and optimizes using a code inner product for similarity-preserving learning. Adversarial Binary Coding (ABC)~\cite{liu2019adversarial} uses adversarial learning to optimize between binary and real-valued features using a Wasserstein loss~\cite{frogner2015learning}. In~\cite{wang2020faster}, a 2048-dimensional hashing code is learned using self-distillation across at different stages of the network. DeepSSH~\cite{zhao2018deepssh} uses attribute- and identity-level hash codes using a sigmoid cross-entropy loss as a relaxation of the sign function. In contrast, our approach employs the soft-sign activation, facilitating better convergence due to a gradual gradient slope and tighter bounds enforced on the gradient values (similar to label smoothing).

\section{Methodology}
\label{sec:methodology}

\subsection{Preliminaries}
The whole-body (person) images and corresponding identity labels from the training set are denoted as $X_{train}=\{x^{t}_1, x^{t}_2, \dots, x^{t}_n\}$ and $Y_{train}=\{y^{t}_1, y^{t}_2, \dots, y^{t}_n\}$, respectively, where $n$ is the total number of images in the set. The aim is to learn a discriminative hash code function $\phi(x;\theta)$ that can match images from disjoint sets $X_{query}$ and $X_{gallery}$, where $\theta$ denotes trainable parameters of the network. To compare hash (or binary) representations, we use the Hamming distance metric.
The representations are directly optimized (such as in the triplet loss) as well as fed to a classifier to optimize the logits of the classifier.  Figure~\ref{fig:methodology} captures the overall architecture of our proposed network. 

\subsection{Hash-based Dynamic Network}
To generate a discriminative binary code representation of the person image, we utilize the ResNet50~\cite{he2016deep} architecture as the encoder to fairly compare with other methods. However, the encoder can be easily swapped to save further computation. To be able to halt computation at different computational budgets, we place $n_h$ hash exit blocks across the network, roughly equally spaced according to computational cost. In our current implementation, we have set $n_h$ to 4, but it can be easily modified to meet the requirements of the backbone network or application.

The network comprises four stages producing a 256-, 512-, 1024-, and 2048-dimensional feature, respectively.

Given an image $x_i^t$ from the training set, the network produces four representations at different stages, denoted as:
\begin{equation}
\label{eq:stages}
\begin{aligned}
    s_1 = \phi^{stage1}(x_i^t) ,\hspace{1cm}
    s_2 = \phi^{stage2}(s_1), \\
    s_3 = \phi^{stage3}(s_2) ,\hspace{1cm}
    s_4 = \phi^{stage4}(s_3).
\end{aligned}
\end{equation}
The initial layer ($s_1$ in Eq.~\ref{eq:stages}) contain more local and fine-level details, and the spatial dimensions retain most of the discriminative information. For example, at $s_1$, the input image with size $256 \times 128 \times 3$ (H $\times$ W $\times$ C) is mapped to a $64 \times 32 \times 256$ (H $\times$ W $\times$ C) feature representation. Global pooling on the spatial dimensions loses too much information,  resulting in poor performance. Therefore, we utilize a part-based local pooling operation that retains spatial information. Specifically, we identify that person images can be categorized into four parts (head, upper torso, lower torso, and feet). We split the feature map into four spatial parts in the height dimension, producing four $16 \times 32 \times 256$ dimensional-tensors. 
These are then average pooled together to generate a $1 \times 1 \times 256$ feature. Mathematically, it is:
\begin{equation}
\label{eq:s1hat}
    \hat{s}_1 = concat(Avg(s_1\{i:i+16, 32, 256\})), 
\end{equation}
where $i \in \{0,16,32, 48\}$ and $\hat{s}_1$ denotes that part-based pooling applied on $s_1$. 

To generate the hash code, we attach a novel Hash-Exit (HE) block to each of the intermediate representations ($s_2$, $s_3$, $s_4$ from Eq.~\ref{eq:stages}, and $s_1$ from Eq.~\ref{eq:s1hat}). 
The first two blocks, shown in  Eq.~\ref{eq:stage12}, serve the dual purpose of bridging the gap between fine and coarse-level features in the earlier layers and generating a concise hash code. 
Specifically, the block consists of using a fully connected layer ($FC$) with a batch normalization layer ($BN$) and a hyperbolic tangent ($tanh$) function, followed by another $FC$ layer to introduce non-linearity. The hash code is generated by first centering the features around 0 using a batch normalization ($BN$) layer and then finally using a $sign$ function. 
Formally, it is:
\begin{equation}
\label{eq:stage12}
\begin{aligned}
    HE_{1, 2} = FC_{512}-BN_{512}-tanh-\\\underbrace{FC_{256}-BN_{256}-sign}_{hash}, \\
\end{aligned}
\end{equation}
where the subscript for $FC$ and $BN$ denotes the output dimension, and the $sign$ function returns 1 if output value is greater than 0 and returns -1 otherwise.

The third and fourth (final) stage feature representation are generic enough to not need a fine-to-coarse transformation, and hence the representations are directly fine-tuned without using a non-linear transformation (shown in Eq.~\ref{eq:stage34}). 
\begin{equation}
\label{eq:stage34}
\begin{aligned}
    HE_{3} = \underbrace{BN_{1024}-sign}_{hash}, 
    HE_{4} = \underbrace{BN_{2048}-sign}_{hash}, \\
\end{aligned}
\end{equation}



We also attach a classifier at the end of each output from the HE blocks for cross-entropy minimization. However, the classifier is discarded after training.

\subsection{Multi-Exit Optimization with Soft Sign}
Given a training set $X_{train}$, we first employ the triplet loss with hard mining that minimizes the distance between the most different representations of the same identity in the batch, and maximizes the distance between two similar representations of distinct identities. For each iteration, P distinct identities with K images per P (identities) are sampled in a batch. Next, for each sample in the batch (anchor), the hardest (furthest) positive and hardest (closest) negative are selected to compute the loss. Mathematically, it can be denoted as:
\begin{equation}
\label{eq:triplet}
\begin{aligned}
    L_{T} = \sum_{^P} \sum_{^K} max(
    \overbrace{d(\phi(x^a), \phi(x^p))}^\text{hardest positive}\\
    - \underbrace{d(\phi(x^a), \phi(x^n))}_\text{hardest negative}
    + margin, 
    0)
\end{aligned}
\end{equation}
where $x^a$, $x^p$, and $x^n$ denote the anchor, positive, and negative samples, respectively, and margin is the minimum gap set to 0.2 as in~\cite{hermans2017defense}.
To bridge the gap between hash codes and the embedding representation, we employ the negative log likelihood loss on the classifier logits: 
\begin{equation}
    L_{C} = - \sum_{i=0}^{N} y_i log(\hat{y}_i),
\end{equation}
where $y_i$ is the true label of the identity and $\hat{y_i}$ is the softmax probability of the class. This ensures that the original representation is discriminative enough, whereas the classification probability of the hash codes aligns with the features. 
Lastly, to minimize the distance and keep it differentiable~\cite{6247912}, we minimize the inner product:

\begin{equation}
    L_{R} = \mathbb{E}(feat_{ori} \cdot feat_{ori}^{T} - feat_{hash} \cdot feat_{hash}^{T})^2,
\end{equation}
between the distance matrix of the original continuous features ($feat_{ori}$) and the hash features ($feat_{hash}$) for the top-5 ranks in the batch.

This ensures we are able to optimize the distance between continuous and hash features, without considering separability of all samples that might hinder the learning. 

The final loss is denoted as:
\begin{equation}
    \begin{aligned}
    L_{final} = \lambda_{1} L_{T}(feat_{ori}) + \lambda_{2}L_{C}(feat_{hash})\\ + \lambda_{3} L_{R}(feat_{ori}, feat_{hash})
    \end{aligned}
\end{equation}
where $\lambda_{1}$, $\lambda_{2}$, and $\lambda_{3}$ are used to balance the losses. The empirically determined values are specified in Section~\ref{sec:implementation}.
Note that the Hamming distance and the sign activation function is not differentiable, and hence a smooth sign function and the inner-product distance~\cite{6247912} is used to make it convex and tractable. 

\subsection{Exit using Training Statistics (ETS)}
\label{sec:exit_policy}
To support adaptive inference, we need to determine when samples can exit the network---meaning the most effective HE block in terms of both efficiency and discriminability. Previous work uses classifier confidence, but in a retrieval problem, we have unknown number of classes, making this infeasible. Heuristics approaches such as similarity metric or margin between have been explored~\cite{yang2020resolution}, but this needs hand-tuning of the distance thresholds and is not generalizable. In this work, we propose to predict which samples are easy by using a temporal network that predicts whether or not to exit early. The temporal network accepts the $\hat{s_{1}}$ representation of the query and the top-4 matches as input, and consists of a gated recurrent unit (GRU) with two hidden stages, followed by a ReLU activation and a three-output classifier with outputs $easy, skip,$ and $hard$.
However, the challenge is to train the network to classify such samples without fine-tuning on the test set. To address this, we utilize our training phase to collect statistics of the number of flips in the top-1 retrieval for the sample. Specifically, assuming that we train our model for 100 epochs, we collect the top-1 decisions of the training samples at every 10 epochs. An example for an easy/hard sample is denoted as:
\begin{equation}
\label{eq:checkmarks}
\begin{aligned}
    T_{easy} = [\times, \underbrace{\times, \checkmark
}_{flip}, \checkmark, \checkmark, \checkmark, \checkmark, \checkmark, \checkmark, \checkmark] \\
    T_{hard} = [\times, \overbrace{\times, \checkmark,}^{flip} \underbrace{\checkmark, \times,}_{flip} \times, \overbrace{\times, \checkmark,}^{flip} \checkmark, \checkmark]
\end{aligned}
\end{equation}
where $\checkmark$ and $\times$ denotes a correct and incorrect prediction, respectively.
At the end of the training phase, if the unit flips its decision greater than 2 times, this means that the sample is $hard$ to predict ($T_{hard}$ in Eq.~\ref{eq:checkmarks}). For samples that have large number of flips ($>6$), are usually noise or low resolution samples. For such samples, we classify it as $skip$ and exit early as we are not sure the network will consistently classify it incorrectly. For the \textit{hard} samples, we use a gallery relation heuristic (Gallery Separability) where the top-2 retrievals belonging to different identities are compared, and if the distance between these retrievals are greater than a threshold, we exit early as the top retrieval is well-separated from the other samples.

\section{Experimental Analysis}
\subsection{Datasets}
We employ three datasets for benchmarking: Market1501~\cite{zheng2015scalable}, MSMT17 (Multi-Scene Multi-Time 17)~\cite{wei2018person}, and BGC1 (BRIAR Government Collection 1)~\cite{cornett2023expanding} dataset. 

Market1501~\cite{zheng2015scalable} captures imagery from six
cameras in the Tsinghua University campus, consisting of 12,936 training images belonging to 751 identities
and 19,732 testing images of 750 identities. The annotation process using the Deformable Part Model~\cite{felzenszwalb2009object} introduces improper crops, making it a useful dataset that reflects realistic scenarios.

MSMT17~\cite{wei2018person} is a large
scale dataset consisting of 4,101 identities
captured by 12 outdoor and 3 indoor cameras, across varying time scales and scenes. It is split into 32,621 training, 11,659 query and 82,161 gallery images.

BGC1~\cite{cornett2023expanding} is a new dataset comprising of unconstrained face and whole-body images and videos from close-range, 100m, 200m, 400m, and 500m distance ranges for 150 identities. In this work, we use 45,111 images of 126 total identities corresponding to close-range and 200m images for the training set. A related sequestered dataset called BTS1 (BRIAR Test Set) is used to test our model comprising of 15,666 images of 67 identities. 

\subsection{Implementation Details}
\label{sec:implementation}
All experiments are run on a Intel Core i7 CPU equipped with a single NVIDIA GeForce RTX 2080 Ti with 11GB memory. The input images are resized to $256 \times 128$. We use random flipping and random erasing with a probability of 0.5. The batch size is 64, consisting of 16 distinct identities with 4 samples per identity. The backbone network is initialized with ImageNet~\cite{deng2009imagenet} weights. The lambda values are set to $\lambda_1=0.35, \lambda_2=1, \text{and}~\lambda_3=100$, evaluated using cross-validation data. The method is trained for 100 epochs with learning rate set to 3e-4, and decayed gradually to 3e-5 and 3e-6
by 40 and 70 epochs. For complete reproducibility, code will be made available in PyTorch~\cite{NEURIPS2019_9015}.

\begin{table*}[h!]
    \caption{Market1501 performance. \textbf{\textcolor{blue}{B}} denotes binary-valued representation whereas \textbf{\textcolor{orange}{C}} denotes continuous-valued representation. }
    \label{tab:market1501}
    \centering
    \resizebox{1.7\columnwidth}{!}{
    
    \begin{tabular}{llllllll}
     
    \cline{1-8}
     \multicolumn{4}{c}{\textbf{\underline{Hash-based Methods}}} & \vline & \multicolumn{3}{l}{\textbf{\underline{SOTA ReID}}}  \\
        \cline{1-8}

 Method & Type/Length & R-1 & mAP &  \vline~Method & Type/Length & R-1 (S4) & mAP \\
    
        \hline


   HashNet~\cite{Cao_2017_ICCV} & \textbf{\textcolor{blue}{B}}/512 & 29.20 &  19.10 & \vline~PNGAN~\cite{qian2018pose} & \textbf{\textbf{\textcolor{orange}{C}}}/2048 & 89.40 & 72.60 \\
     DeepSSH~\cite{zhao2018deepssh} & \textbf{\textcolor{blue}{B}}/512 & 46.50 & 24.10 & \vline~SVDNet~\cite{sun2017svdnet} & \textbf{\textcolor{orange}{C}}/2048  & 82.30 & 61.10 \\
    ABC~\cite{liu2019adversarial} & \textbf{\textcolor{blue}{B}}/2048 & 81.40 & 64.70 & \vline~OSNet~\cite{zhou2019omni} & \textbf{\textcolor{orange}{C}}/2048 & 94.20 & 82.60 \\
    DCH~\cite{cao2018deep} & \textbf{\textcolor{blue}{B}}/512 & 40.70  & 20.20 & \vline~BOT~\cite{luo2019bag} & \textbf{\textcolor{orange}{C}}/2048  & 94.50 & 85.90  \\
    CtF~\cite{wang2020faster} & \textbf{\textcolor{blue}{B}}/2048 & 93.70  & 84.10 & \vline~MLFN~\cite{chang2018multi} & \textbf{\textcolor{orange}{C}}/2048 & 90.10 & 74.30  \\
    PDH~\cite{zhu2017part} & \textbf{\textcolor{blue}{B}}/512 & 44.60 & 24.30 & \vline~HaCNN~\cite{li2018harmonious} & \textbf{\textcolor{orange}{C}}/512 & 90.90 & 75.60   \\
    DSRH~\cite{zhao2015deep} & \textbf{\textcolor{blue}{B}}/512 & 27.10 & 17.70 & \vline~TriNet~\cite{hermans2017defense} & \textbf{\textcolor{orange}{C}}/2048 & 84.90 & 69.10  \\
    OURS-HE & \textbf{\textcolor{blue}{B}}/2048 & \textbf{94.18} & \textbf{84.85} &  \vline~OURS-HE & \textbf{\textcolor{blue}{B}}/2048 &\textbf{94.18} &\textbf{84.85}  \\
    \cmidrule{1-8}
     \multicolumn{8}{c}{\textbf{\underline{Dynamic Networks}}} \\
    \cmidrule{1-8}
    
      Method & Type & Length & R-1 (S1) & R-1 (S2) & R-1 (S3) & R-1 (S4) & mAP \\
    \cmidrule{1-8}
    BranchyNet~\cite{teerapittayanon2016branchynet}& \textbf{\textcolor{orange}{C}} & 14880/13440/2048 & 38.81 & 58.05 & - & 80.05 & 62.76 \\
     
    MSDNet~\cite{huang2017multi} & \textbf{\textcolor{orange}{C}} & 384/384/352/204 & 58.79 & 61.67 & 64.01  & 63.51 & 38.07 \\
    RANet~\cite{yang2020resolution} & \textbf{{\textcolor{orange}{C}}} & 576/1088/641/897 & 58.55 & 59.47 & 65.26 & 65.05  & 40.19 \\
    DaRE(R)~\cite{wang2018resource} & \textbf{{\textcolor{orange}{C}}} & 128/128/128/128  & 62.86 & 74.20 & 82.30  & 83.91  & 65.40 \\
    DaRE+RE(R)~\cite{wang2018resource} & \textbf{\textcolor{orange}{C}} & 128/128/128/128  & 62.47 & 78.38 & 87.05   & 87.77  & 74.34 \\
    OURS+HE & \textbf{\textcolor{blue}{B}} & 256/256/1024/2048 & \textbf{71.29} & \textbf{79.93} & \textbf{92.10} & \textbf{92.10} & \textbf{82.08} \\
    \hline
    \end{tabular} 
    }
\end{table*}

\begin{table*}[h!]
    \caption{MSMT17 Performance. \textbf{\textcolor{blue}{B}} denotes binary-valued representation whereas \textbf{\textcolor{orange}{C}} denotes continuous-valued representation. }
    \centering
    \resizebox{1.7\columnwidth}{!}{
    
    \begin{tabular}{llllllll}
     
    \cline{1-8}
     \multicolumn{4}{c}{\textbf{\underline{Hash-based Methods}}} & \vline & \multicolumn{3}{l}{\textbf{\underline{SOTA ReID}}}  \\
        \cline{1-8}

 Method & Type/Length & R-1 & mAP &  \vline~Method & Type/Length & R-1 (S4) & mAP \\
    
        \hline


   HashNet~\cite{Cao_2017_ICCV} & \textbf{\textcolor{blue}{B}}/512 & 23.55 &  10.65 & \vline~PCB~\cite{sun2018beyond} & \textbf{\textcolor{orange}{C}}/2048 & 68.20 & 40.40  \\
     DTSH~\cite{wang2017deep} & \textbf{\textcolor{blue}{B}}/512 & 47.37  & 25.61 & \vline~GLAD~\cite{wei2017glad} & \textbf{\textcolor{orange}{C}}/2048  & 61.40 & 34.00 \\
    ABC~\cite{liu2019adversarial} & \textbf{\textcolor{blue}{B}}/2048 & - & - & \vline~OSNet~\cite{zhou2019omni} & \textbf{\textcolor{orange}{C}}/2048 & 79.10 & 55.10 \\
    QSMI~\cite{PASSALIS2021116146} & \textbf{\textcolor{blue}{B}}/512 & 16.21  & 9.88 & \vline~IANet~\cite{hou2019interaction} & \textbf{\textcolor{orange}{C}}/2048  & 75.50 & 46.80  \\
    CtF~\cite{wang2020faster} & \textbf{\textcolor{blue}{B}}/2048 & 75.95  & 51.36 & \vline~MLFN~\cite{chang2018multi} & \textbf{\textcolor{orange}{C}}/2048 & 66.40 & 37.20 \\
    PDH~\cite{zhu2017part} & \textbf{\textcolor{blue}{B}}/512 & 37.13 & 16.90 & \vline~HaCNN~\cite{li2018harmonious} & \textbf{\textcolor{orange}{C}}/512 & 64.70 & 37.20   \\
    DSRH~\cite{zhao2015deep} & \textbf{\textcolor{blue}{B}}/512 & 29.91 & 14.75 & \vline~DGNet~\cite{zheng2019joint} & \textbf{\textcolor{orange}{C}}/2048 & 77.20 & 52.30  \\
    OURS-HE & \textbf{\textcolor{blue}{B}}/2048 & \textbf{76.81} & \textbf{51.41} &  \vline~OURS-HE & \textbf{\textcolor{blue}{B}}/2048 & \textbf{76.81} & \textbf{51.41}  \\
    \cmidrule{1-8}
     \multicolumn{8}{c}{\textbf{\underline{Dynamic Networks}}} \\
    \cmidrule{1-8}
    
      Method & Type & Length & R-1 (S1) & R-1 (S2) & R-1 (S3) & R-1 (S4) & mAP \\
    \cmidrule{1-8}
    BranchyNet~\cite{teerapittayanon2016branchynet}& \textbf{\textcolor{orange}{C}} & 14880/13440/2048 & 9.17 & 21.27 & - & 49.93 & 27.51 \\
     
    MSDNet~\cite{huang2017multi} & \textbf{\textcolor{orange}{C}} & 384/384/352/304 & 15.56 & 20.15 & 20.98 & 21.94 & 8.53 \\
    RANet~\cite{yang2020resolution} & \textbf{\textcolor{orange}{C}} & 576/1088/641/897 & 13.47 & 14.90 & 19.99 & 20.44  & 8.04 \\
    DaRE(R)~\cite{wang2018resource} & \textbf{\textcolor{orange}{C}} & 128/128/128/128  & 16.81 & 19.56 & 51.76  & 52.11  & 30.01 \\
    DaRE+RE(R)~\cite{wang2018resource} & \textbf{\textcolor{orange}{C}} & 128/128/128/128   & 15.49 & 19.92 & 53.18  & 54.77  & 30.31 \\
    OURS+HE & \textbf{\textcolor{blue}{B}} & 256/256/1024/2048 & \textbf{28.91} & \textbf{47.74} & \textbf{71.31}  & \textbf{71.52}  & \textbf{46.66} \\
    \hline
    \end{tabular} 
   }
    \label{tab:msmt17}
\end{table*}

\subsection{Quantitative Results}
Table~\ref{tab:market1501} and Table~\ref{tab:msmt17} presents comparisons on the Market1501 and MSMT17 datasets with ReID-specific, Hash-based, and Dynamic Networks. We present two versions of our method: Ours - HE denotes without exit blocks and Ours + HE denotes the version with exit blocks. From Table~\ref{tab:market1501}, our method achieves slightly improved performance over CtF. While ReID-specific methods such as OSNet and BOT perform marginally better than our method, it is important to note that we use hash/binary features, which significantly improves searching performance. Compared to dynamic networks, our method achieves 8.82\%, 1.55\%, 5.05\% and 4.33\% improvement in rank-1 accuracy over the four stages, respectively. Moreover, the mAP score improves from 74.34\% to 82.08\% over DaRE. 

\begin{figure}[h]
    \centering
    \includegraphics[width=0.85\columnwidth]{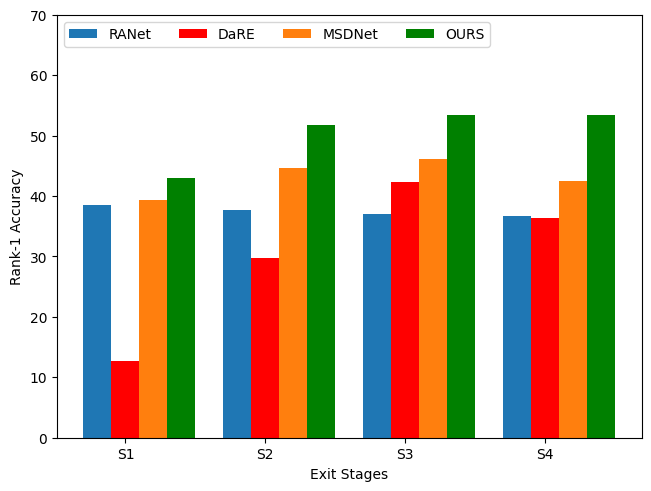}
    \caption{Performance on the BGC1 Dataset.}
    \label{fig:bgcperformance}
\end{figure}

Similar trends can be seen on the MSMT17 dataset in Table~\ref{tab:msmt17}, with 12.10\%, 26.47\%, 18.13\%, and 16.75\% over the four stages, respectively. The mAP score increases by 16.35\% over the previous best-performing method. 
Figure~\ref{fig:bgcperformance} showcases the performance on the BGC1 dataset compared with related dynamic networks.

\subsection{Bit Length}
In Table~\ref{tab:bitlength}, we present the performance trend on various hash-bit lengths with the \textit{OURS-HE} model. The final column indicates the time required in $10^{-6}$ seconds for computing the Hamming distance between hash features compared to the euclidean distance for continuous-valued features. For a more detailed comparison, we refer readers to~\cite{wang2020faster}.  
Even with a relatively short length of 128 bits, the model achieve 84.26\% rank-1 accuracy. However, the correct match separability is low as observed by the mAP score. 
With increase in length, there is a significant improvement in the mAP scores, while the rank-1 accuracy improves gradually. The binary representation significantly performs better ($>$90x), demonstrating the benefit of hash features. As observed in~\cite{wang2020faster}, the query search time (in seconds) on the Market1501 dataset is $2.2$ for the 2048-dimensional continuous-valued representation and $2.8 \times 10^{-1}$ for the 2048-dimensional hash-valued representation. In our approach, we employ a 256-dimensional feature for the first two exit blocks, and 1024- and 2048-dimensional hash feature for the last two exit blocks, respectively. 
As most ($>70\%$) samples exit early at stage 1 with 256-dimensional features, 80\% of the networks computational cost is saved, while reducing the total query search time to $1.1 \times 10^{-1}$. 
This leads to an improvement of 60\% over using only 2048-dimensional hash codes. 

\begin{table}[htb]
    \centering
    \begin{tabular}{|c|c|c|c|}
    \hline
         Length & Rank-1 & mAP & Time (B:C) in $10^{-6}$ s.  \\
    \hline
         128 & 84.26 & 67.22 & 2.8:260 (92x) \\
         256 & 89.99 & 75.54 & 3.3:500 (151x) \\
         512 & 91.12 & 79.06 & 4.4:1000 (227x) \\
         1024 & 91.80 & 80.63 & 7.1:2000 (281x) \\
         2048 & 94.18 & 84.85 & 17:3900 (229x) \\
    \hline
    \end{tabular}
    \caption{Longer codes increases comparison time but improves  separability as seen in the mAP scores. }
    \label{tab:bitlength}
\end{table}

\subsection{Budgeted Inference}
\label{sec:budgeted_inference}
\begin{figure}[h]
    \centering
    \includegraphics[width=0.8\columnwidth]{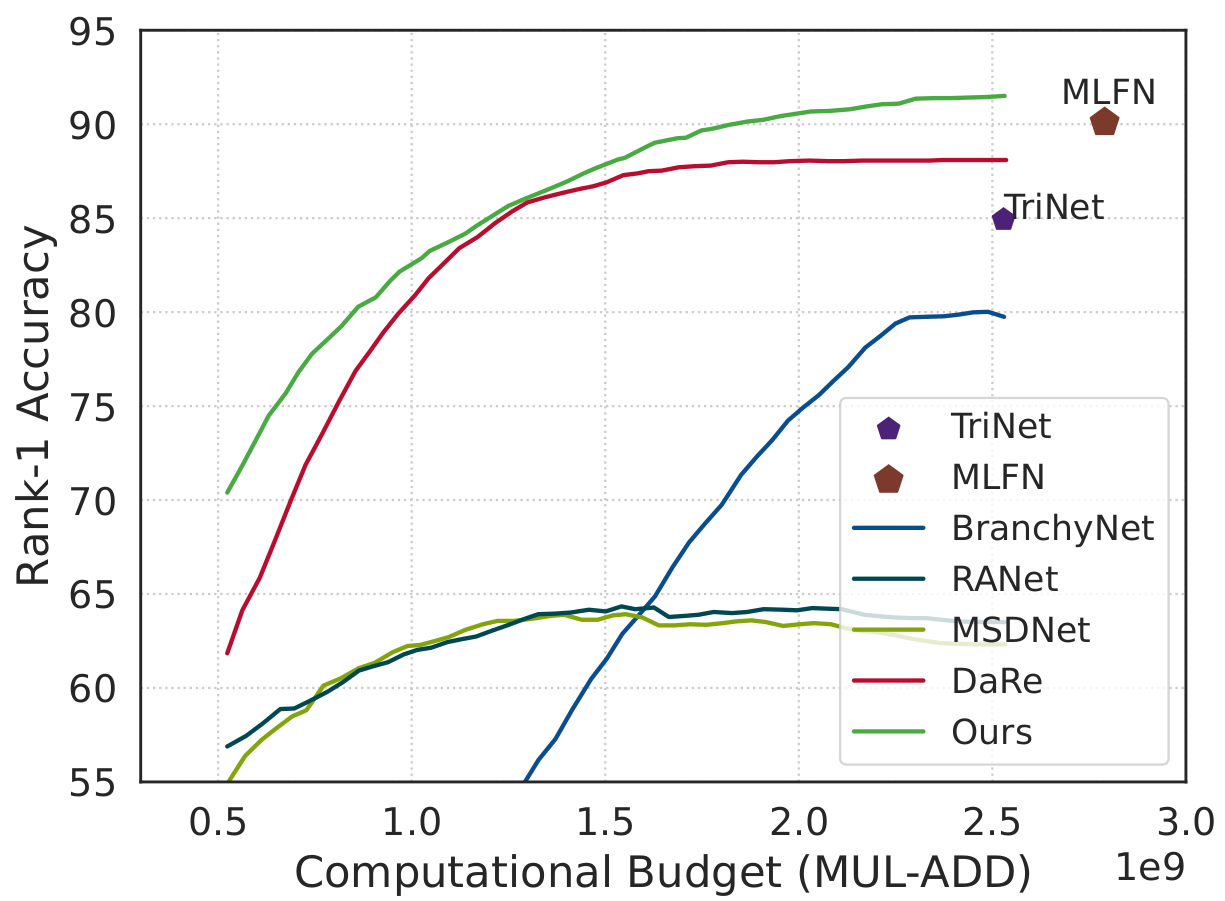}
    \caption{Budgeted inference compared with various dynamic networks. * denotes performance of non-dynamic networks. }
    \label{fig:budgeted_inference}
\end{figure}

In the budgeted inference setting shown in Figure~\ref{fig:budgeted_inference}, the model operates within a predefined computational budget, represented by the x-axis indicating the number of floating-point operations (FLOPS), to classify all query samples. In this, we employ early-exiting of \textit{easy/skip} samples while propagating \textit{hard} examples. 
We compare with four SOTA dynamic networks: MSDNet~\cite{huang2017multi}, RANet~\cite{yang2020resolution}, BranchyNet~\cite{teerapittayanon2016branchynet}, and DaRE~\cite{wang2018resource}. Additionally, we compare with SOTA ReID methods: TriNet~\cite{hermans2017defense} and MLFN~\cite{chang2018multi}. RANet and MSDNet exhibits low initial performance, with only gradual increase in performance as budget increases. BranchyNet demonstrates a steep increase, indicating that earlier stages have very low performance compared to later stages. Compared to DaRE, our method consistently performs better with substantial performance gap between the lowest and highest budget point. This signifies that throughout the network, the model is able to classify most samples accurately compared to other methods. 
To ensure fairness in comparison, all methods utilize the same exit policy.

\begin{figure}
    \centering
    \includegraphics[width=0.8\columnwidth]{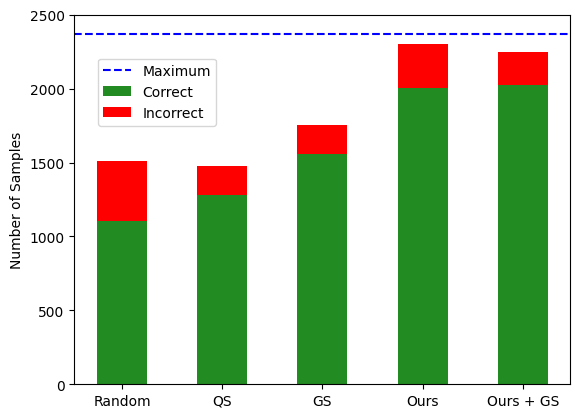}
    \caption{Number of correct  (\textcolor{green}{green bar}) and incorrect exits (\textcolor{red}{red bar}) at stage 1. The maximum dotted line denotes the maximum number of samples that can be correctly exited at stage 1.}
    \label{fig:policy}
\end{figure}

\begin{figure*}[htb]
    \centering
    \includegraphics[width=1.5\columnwidth]{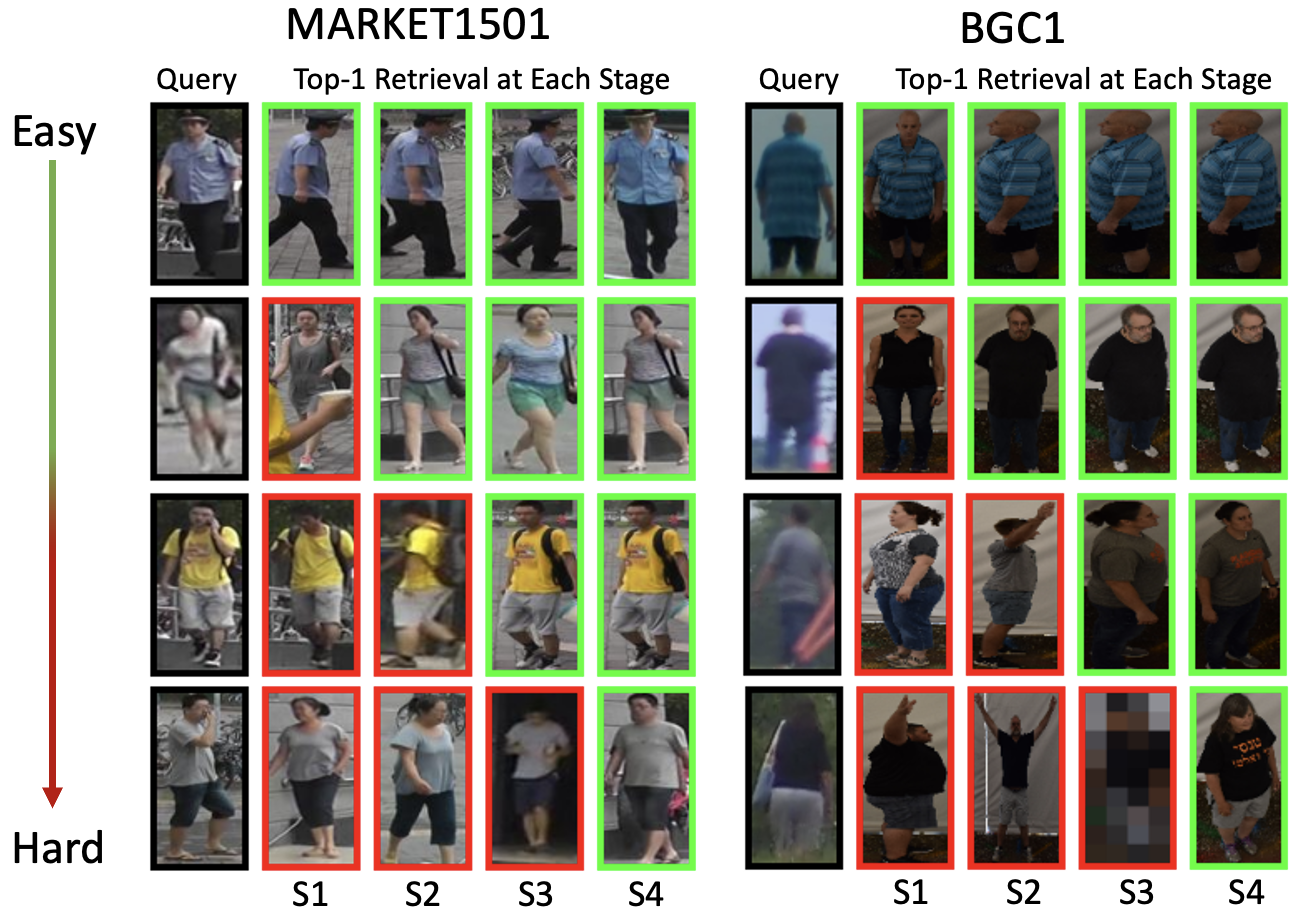}
    \caption{Stage-wise retrieval performance on the Market1501 and BGC1 datasets. \textcolor{green}{Green} border denotes correct retrieval whereas \textcolor{red}{Red} border denotes incorrect retrieval. All subjects consent to image publication. One image (row 4) is pixelated for privacy. }
    \label{fig:stage-performance}
\end{figure*}

\subsection{Earliest Exit Performance}
\label{sec:stage1exit}
In Figure~\ref{fig:policy}, we present the results on five distinct exit policy  techniques for exiting at stage 1 ($HE_{1}$). \textit{Random} exiting involves randomly determining the sample to exit with a 50\% probability. \textit{Query Separability (QS)} is the Hamming distance between the query and the top-1 gallery sample for determining whether to exit. \textit{Gallery Separability (GS)} is the distance between the top-2 matches in the gallery. If the separability exceeds a threshold value, the sample is exited. \textit{Ours} is the proposed GRU-based classifier discussed in Section~\ref{sec:exit_policy}, and \textit{Ours+GS} is the combination of the GRU and GS. The stacked bar represents the count of exited samples, where the green and red bar denote if the sample had the correct or incorrect top-1 retrieval at $HE_{1}$, respectively. Based on the figure, it is evident that the \textit{Random} approach is the least favorable choice, although it still demonstrates acceptable performance due to its strong performance at $HE_{1}$. Both \textit{QS} and \textit{GS} appear to be conservative approaches, resulting in low numbers of correct exits. \textit{Ours} achieves high number of accurate classifications for early-exiting, but also has the most number of incorrect exits. The combination of the GRU classifier and GS heuristic (\textit{Ours+GS}) exhibits the best performance, achieving high number of correct matches and low number of incorrect matches.
The maximum line refers to the highest number of correct exits achievable at $HE_{1}$. All thresholds are determined using cross-validation.

\begin{figure}[htb]
    \centering
    \includegraphics[width=\columnwidth]{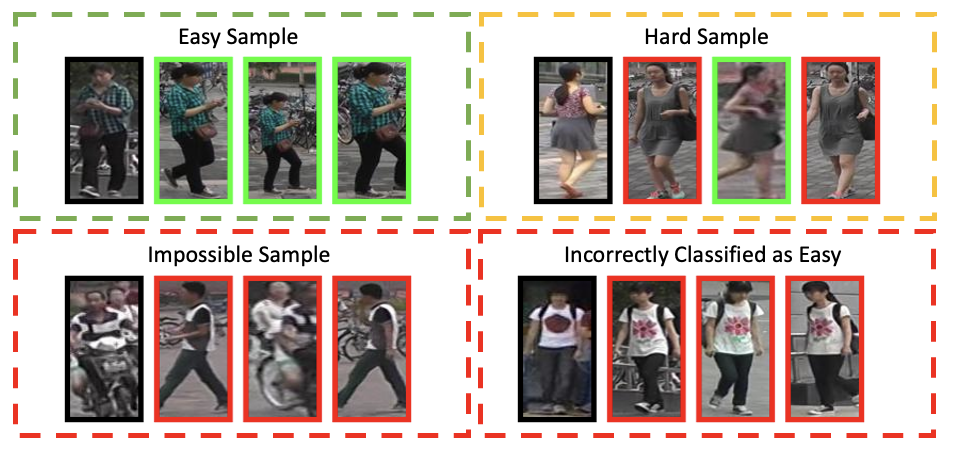}
    \caption{Qualitative performance using the ETS policy. \textcolor{green}{Green} border denotes correct retrieval whereas \textcolor{red}{red} border denotes incorrect retrieval for a given query image (\textbf{black} border). }
    \label{fig:gru_qual}
\end{figure}

\subsection{Qualitative Results}
\label{sec:qualitative_results}
Figure~\ref{fig:stage-performance} illustrates the top match obtained at each stage, with different levels of difficulty of query samples. In the first row is an easy sample, where the uniform (and distinct patterns) facilitates clear separation from other identities, resulting in correct retrievals at every exit stage. The second row is a low-resolution query input, which  benefits from additional computation to accurately retrieve the corresponding gallery match. The third and fourth row represent hard query samples, where the similarity in clothing negatively impacts performance at earlier stages. However, with full computation and longer hash codes, the model is eventually able to correctly classify these challenging samples.

Figure~\ref{fig:gru_qual} shows results using the ETS policy. The \textit{Easy Sample} category stands out from other identities due to distinct patterns and colors, such as a checkered shirt. \textit{Skip/Impossible samples} typically consist of noisy or ambiguous images where multiple identities appear together. \textit{Hard Samples} refer to individuals wearing similar clothing and in similar views, requiring additional computation to generate reliable cross-view representations. Finally, incorrect classifications from the classifier are presented, where identities wearing similar clothing patterns and accessories (e.g., backpack) in similar poses result in erroneous matches. To reduce these mis-classifications, a combination of a heuristic and our GRU classifier is employed, leading to reduced errors as demonstrated in Figure~\ref{fig:policy}.

\section{Conclusion}
\label{sec:conclusion}
In conclusion, this work introduces a novel hash-based dynamic network capable of adapting its computation based on the difficulty of input samples. We leverage a GRU-based ETS policy to assess the complexity, considering both the query and top-gallery samples to make informed decisions regarding early exits. To ensure discriminability between hash and continuous-valued features, we incorporate a ranking regularization technique that optimizes feature similarity.
The adoption of hash representation results in a significant improvement in query-gallery matching compared to continuous-valued representations, while saving computational costs because of the dynamic capability of the network. Our work establishes a robust baseline for an input-adaptive hash network for biometric applications.

\section{Acknowledgement}
This research is based upon work supported in part by DEVCOM Army Research Laboratory (ARL) under contract W911NF-21-2-0076, DEVCOM ARL and the National Strategic Research Institute (NSRI) under contract FA4600-20-D-0003, and the
Office of the Director of National Intelligence (ODNI), Intelligence Advanced Research Projects Activity (IARPA), via 2022-21102100002. The views and conclusions contained herein are those of the authors and should not be interpreted as necessarily representing the official policies, either expressed or implied, of DEVCOM ARL, NSRI, ODNI, IARPA, or the U.S.~Government. The U.S.~Government is authorized to reproduce and distribute reprints for governmental purposes not withstanding any copyright annotation therein.

{\small
\bibliographystyle{ieee_fullname}
\bibliography{nikhalriggan_wacv2024}

\begin{thebibliography}{10}\itemsep=-1pt

\bibitem{cao2018deep}
Yue Cao, Mingsheng Long, Bin Liu, and Jianmin Wang.
\newblock Deep cauchy hashing for hamming space retrieval.
\newblock In {\em Proceedings of the IEEE Conference on Computer Vision and
  Pattern Recognition}, pages 1229--1237, 2018.

\bibitem{Cao_2017_ICCV}
Zhangjie Cao, Mingsheng Long, Jianmin Wang, and Philip~S. Yu.
\newblock Hashnet: Deep learning to hash by continuation.
\newblock In {\em Proceedings of the IEEE International Conference on Computer
  Vision (ICCV)}, Oct 2017.

\bibitem{chang2018multi}
Xiaobin Chang, Timothy~M Hospedales, and Tao Xiang.
\newblock Multi-level factorisation net for person re-identification.
\newblock In {\em Proceedings of the IEEE conference on computer vision and
  pattern recognition}, pages 2109--2118, 2018.

\bibitem{cornett2023expanding}
David Cornett, Joel Brogan, Nell Barber, Deniz Aykac, Seth Baird, Nicholas
  Burchfield, Carl Dukes, Andrew Duncan, Regina Ferrell, Jim Goddard, et~al.
\newblock Expanding accurate person recognition to new altitudes and ranges:
  The briar dataset.
\newblock In {\em Proceedings of the IEEE/CVF Winter Conference on Applications
  of Computer Vision}, pages 593--602, 2023.

\bibitem{deng2009imagenet}
Jia Deng, Wei Dong, Richard Socher, Li-Jia Li, Kai Li, and Li Fei-Fei.
\newblock Imagenet: A large-scale hierarchical image database.
\newblock In {\em 2009 IEEE conference on computer vision and pattern
  recognition}, pages 248--255. Ieee, 2009.

\bibitem{felzenszwalb2009object}
Pedro~F Felzenszwalb, Ross~B Girshick, David McAllester, and Deva Ramanan.
\newblock Object detection with discriminatively trained part-based models.
\newblock {\em IEEE transactions on pattern analysis and machine intelligence},
  32(9):1627--1645, 2009.

\bibitem{frogner2015learning}
Charlie Frogner, Chiyuan Zhang, Hossein Mobahi, Mauricio Araya, and Tomaso~A
  Poggio.
\newblock Learning with a wasserstein loss.
\newblock {\em Advances in neural information processing systems}, 28, 2015.

\bibitem{he2016deep}
Kaiming He, Xiangyu Zhang, Shaoqing Ren, and Jian Sun.
\newblock Deep residual learning for image recognition.
\newblock In {\em Proceedings of the IEEE conference on computer vision and
  pattern recognition}, pages 770--778, 2016.

\bibitem{hector2021scalable}
Rory Hector, Muhammad Umar, Asif Mehmood, Zhu Li, and Shuvra Bhattacharyya.
\newblock Scalable object detection for edge cloud environments.
\newblock {\em Frontiers in Sustainable Cities}, 3:675889, 2021.

\bibitem{hermans2017defense}
Alexander Hermans, Lucas Beyer, and Bastian Leibe.
\newblock In defense of the triplet loss for person re-identification.
\newblock {\em arXiv preprint arXiv:1703.07737}, 2017.

\bibitem{hou2019interaction}
Ruibing Hou, Bingpeng Ma, Hong Chang, Xinqian Gu, Shiguang Shan, and Xilin
  Chen.
\newblock Interaction-and-aggregation network for person re-identification.
\newblock In {\em Proceedings of the IEEE/CVF conference on computer vision and
  pattern recognition}, pages 9317--9326, 2019.

\bibitem{howard2017mobilenets}
Andrew~G Howard, Menglong Zhu, Bo Chen, Dmitry Kalenichenko, Weijun Wang,
  Tobias Weyand, Marco Andreetto, and Hartwig Adam.
\newblock Mobilenets: Efficient convolutional neural networks for mobile vision
  applications.
\newblock {\em arXiv preprint arXiv:1704.04861}, 2017.

\bibitem{huang2017multi}
Gao Huang, Danlu Chen, Tianhong Li, Felix Wu, Laurens Van Der~Maaten, and
  Kilian~Q Weinberger.
\newblock Multi-scale dense networks for resource efficient image
  classification.
\newblock {\em arXiv preprint arXiv:1703.09844}, 2017.

\bibitem{iandola2016squeezenet}
Forrest~N Iandola, Song Han, Matthew~W Moskewicz, Khalid Ashraf, William~J
  Dally, and Kurt Keutzer.
\newblock Squeezenet: Alexnet-level accuracy with 50x fewer parameters and< 0.5
  mb model size.
\newblock {\em arXiv preprint arXiv:1602.07360}, 2016.

\bibitem{li2018harmonious}
Wei Li, Xiatian Zhu, and Shaogang Gong.
\newblock Harmonious attention network for person re-identification.
\newblock In {\em Proceedings of the IEEE conference on computer vision and
  pattern recognition}, pages 2285--2294, 2018.

\bibitem{6247912}
Wei Liu, Jun Wang, Rongrong Ji, Yu-Gang Jiang, and Shih-Fu Chang.
\newblock Supervised hashing with kernels.
\newblock In {\em 2012 IEEE Conference on Computer Vision and Pattern
  Recognition}, pages 2074--2081, 2012.

\bibitem{liu2019adversarial}
Zheng Liu, Jie Qin, Annan Li, Yunhong Wang, and Luc Van~Gool.
\newblock Adversarial binary coding for efficient person re-identification.
\newblock In {\em 2019 IEEE International Conference on Multimedia and Expo
  (ICME)}, pages 700--705. IEEE, 2019.

\bibitem{luo2019bag}
Hao Luo, Youzhi Gu, Xingyu Liao, Shenqi Lai, and Wei Jiang.
\newblock Bag of tricks and a strong baseline for deep person
  re-identification.
\newblock In {\em Proceedings of the IEEE/CVF conference on computer vision and
  pattern recognition workshops}, pages 0--0, 2019.

\bibitem{PASSALIS2021116146}
Nikolaos Passalis and Anastasios Tefas.
\newblock Deep supervised hashing using quadratic spherical mutual information
  for efficient image retrieval.
\newblock {\em Signal Processing: Image Communication}, 93:116146, 2021.

\bibitem{NEURIPS2019_9015}
Adam Paszke, Sam Gross, Francisco Massa, Adam Lerer, James Bradbury, Gregory
  Chanan, Trevor Killeen, Zeming Lin, Natalia Gimelshein, Luca Antiga, Alban
  Desmaison, Andreas Kopf, Edward Yang, Zachary DeVito, Martin Raison, Alykhan
  Tejani, Sasank Chilamkurthy, Benoit Steiner, Lu Fang, Junjie Bai, and Soumith
  Chintala.
\newblock Pytorch: An imperative style, high-performance deep learning library.
\newblock In {\em Advances in Neural Information Processing Systems 32}, pages
  8024--8035. Curran Associates, Inc., 2019.

\bibitem{qian2018pose}
Xuelin Qian, Yanwei Fu, Tao Xiang, Wenxuan Wang, Jie Qiu, Yang Wu, Yu-Gang
  Jiang, and Xiangyang Xue.
\newblock Pose-normalized image generation for person re-identification.
\newblock In {\em Proceedings of the European conference on computer vision
  (ECCV)}, pages 650--667, 2018.

\bibitem{sun2017svdnet}
Yifan Sun, Liang Zheng, Weijian Deng, and Shengjin Wang.
\newblock Svdnet for pedestrian retrieval.
\newblock In {\em Proceedings of the IEEE international conference on computer
  vision}, pages 3800--3808, 2017.

\bibitem{sun2018beyond}
Yifan Sun, Liang Zheng, Yi Yang, Qi Tian, and Shengjin Wang.
\newblock Beyond part models: Person retrieval with refined part pooling (and a
  strong convolutional baseline).
\newblock In {\em Proceedings of the European conference on computer vision
  (ECCV)}, pages 480--496, 2018.

\bibitem{tan2019efficientnet}
Mingxing Tan and Quoc Le.
\newblock Efficientnet: Rethinking model scaling for convolutional neural
  networks.
\newblock In {\em International conference on machine learning}, pages
  6105--6114. PMLR, 2019.

\bibitem{teerapittayanon2016branchynet}
Surat Teerapittayanon, Bradley McDanel, and Hsiang-Tsung Kung.
\newblock Branchynet: Fast inference via early exiting from deep neural
  networks.
\newblock In {\em 2016 23rd International Conference on Pattern Recognition
  (ICPR)}, pages 2464--2469. IEEE, 2016.

\bibitem{wang2020faster}
Guan’an Wang, Shaogang Gong, Jian Cheng, and Zengguang Hou.
\newblock Faster person re-identification.
\newblock In {\em Computer Vision--ECCV 2020: 16th European Conference,
  Glasgow, UK, August 23--28, 2020, Proceedings, Part VIII}, pages 275--292.
  Springer, 2020.

\bibitem{wang2017deep}
Xiaofang Wang, Yi Shi, and Kris~M Kitani.
\newblock Deep supervised hashing with triplet labels.
\newblock In {\em Computer Vision--ACCV 2016: 13th Asian Conference on Computer
  Vision, Taipei, Taiwan, November 20-24, 2016, Revised Selected Papers, Part I
  13}, pages 70--84. Springer, 2017.

\bibitem{wang2018skipnet}
Xin Wang, Fisher Yu, Zi-Yi Dou, Trevor Darrell, and Joseph~E Gonzalez.
\newblock Skipnet: Learning dynamic routing in convolutional networks.
\newblock In {\em Proceedings of the European Conference on Computer Vision
  (ECCV)}, pages 409--424, 2018.

\bibitem{wang2018resource}
Yan Wang, Lequn Wang, Yurong You, Xu Zou, Vincent Chen, Serena Li, Gao Huang,
  Bharath Hariharan, and Kilian~Q Weinberger.
\newblock Resource aware person re-identification across multiple resolutions.
\newblock In {\em Proceedings of the IEEE conference on computer vision and
  pattern recognition}, pages 8042--8051, 2018.

\bibitem{wei2018person}
Longhui Wei, Shiliang Zhang, Wen Gao, and Qi Tian.
\newblock Person transfer gan to bridge domain gap for person
  re-identification.
\newblock In {\em Proceedings of the IEEE conference on computer vision and
  pattern recognition}, pages 79--88, 2018.

\bibitem{wei2017glad}
Longhui Wei, Shiliang Zhang, Hantao Yao, Wen Gao, and Qi Tian.
\newblock Glad: Global-local-alignment descriptor for pedestrian retrieval.
\newblock In {\em Proceedings of the 25th ACM international conference on
  Multimedia}, pages 420--428, 2017.

\bibitem{yang2020resolution}
Le Yang, Yizeng Han, Xi Chen, Shiji Song, Jifeng Dai, and Gao Huang.
\newblock Resolution adaptive networks for efficient inference.
\newblock In {\em Proceedings of the IEEE/CVF conference on computer vision and
  pattern recognition}, pages 2369--2378, 2020.

\bibitem{zhang2018shufflenet}
Xiangyu Zhang, Xinyu Zhou, Mengxiao Lin, and Jian Sun.
\newblock Shufflenet: An extremely efficient convolutional neural network for
  mobile devices.
\newblock In {\em Proceedings of the IEEE conference on computer vision and
  pattern recognition}, pages 6848--6856, 2018.

\bibitem{zhao2015deep}
Fang Zhao, Yongzhen Huang, Liang Wang, and Tieniu Tan.
\newblock Deep semantic ranking based hashing for multi-label image retrieval.
\newblock In {\em Proceedings of the IEEE conference on computer vision and
  pattern recognition}, pages 1556--1564, 2015.

\bibitem{zhao2018deepssh}
Ya Zhao, Sihui Luo, Yezhou Yang, and Mingli Song.
\newblock Deepssh: Deep semantic structured hashing for explainable person
  re-identification.
\newblock In {\em 2018 25th IEEE International Conference on Image Processing
  (ICIP)}, pages 1653--1657. IEEE, 2018.

\bibitem{zheng2015scalable}
Liang Zheng, Liyue Shen, Lu Tian, Shengjin Wang, Jingdong Wang, and Qi Tian.
\newblock Scalable person re-identification: A benchmark.
\newblock In {\em Proceedings of the IEEE international conference on computer
  vision}, pages 1116--1124, 2015.

\bibitem{zheng2019joint}
Zhedong Zheng, Xiaodong Yang, Zhiding Yu, Liang Zheng, Yi Yang, and Jan Kautz.
\newblock Joint discriminative and generative learning for person
  re-identification.
\newblock In {\em proceedings of the IEEE/CVF conference on computer vision and
  pattern recognition}, pages 2138--2147, 2019.

\bibitem{zhou2019omni}
Kaiyang Zhou, Yongxin Yang, Andrea Cavallaro, and Tao Xiang.
\newblock Omni-scale feature learning for person re-identification.
\newblock In {\em Proceedings of the IEEE/CVF international conference on
  computer vision}, pages 3702--3712, 2019.

\bibitem{zhou2019elastic}
Yi Zhou, Yue Bai, Shuvra~S Bhattacharyya, and Heikki Huttunen.
\newblock Elastic neural networks for classification.
\newblock In {\em 2019 IEEE International Conference on Artificial Intelligence
  Circuits and Systems (AICAS)}, pages 251--255. IEEE, 2019.

\bibitem{zhu2017part}
Fuqing Zhu, Xiangwei Kong, Liang Zheng, Haiyan Fu, and Qi Tian.
\newblock Part-based deep hashing for large-scale person re-identification.
\newblock {\em IEEE Transactions on Image Processing}, 26(10):4806--4817, 2017.

\end{thebibliography}
}

\end{document}